\documentclass{article}
\usepackage{spconf,amsmath,graphicx}
\usepackage{subfig}
\usepackage{flushend}
\usepackage{epstopdf}
\usepackage{color}
\usepackage[hyphens]{url}

\title{SKETCH-EPITOMES : DISCRIMINATIVELY MINIMALIST REPRESENTATIONS FOR OBJECT CATEGORIES}

\name{Ravi Kiran Sarvadevabhatla,R. Venkatesh Babu}
\address{Video Analytics Lab, SERC,\\Indian Institute of Science, Bangalore, India.\\ ravikiran@ssl.serc.iisc.in,venky@serc.iisc.in}

\begin{document}
\ninept 
\maketitle

\begin{abstract}
Freehand line sketches are an interesting and unique form of visual representation. Typically, such sketches are studied and utilized as an end product of the sketching process. However, we have found it instructive to study the sketches as sequentially accumulated composition of drawing strokes added over time. Studying sketches in this manner has enabled us to create novel sparse yet discriminative sketch-based representations for object categories which we term \textit{category-epitome}s. Our procedure for obtaining these epitomes concurrently provides a natural measure for quantifying the sparseness underlying the original sketch, which we term epitome-score. We construct and analyze \textit{category-epitome}s and epitome-scores for freehand sketches belonging to various object categories. Our analysis provides a  novel  viewpoint for studying the semantic nature of object categories. 
\end{abstract}


\section{Introduction}
\label{sec:intro}
\begin{figure}[t]
\centering
\includegraphics[width=.15\linewidth]{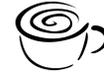}
\caption{In spite of minimal detail, we can recognize the line sketch easily and correctly as belonging to the category \texttt{cup}.}
\label{fig:cup}
\end{figure}

Sketches, as a form of visual representation, exhibit a great variety from realistic portraits to sparsely drawn, stylistic ones. In particular, consider freehand (i.e. hand-drawn) sketches of objects. An instance of such a sketch can be seen in Figure \ref{fig:cup}. Though containing minimal detail, the object category to which it belongs is easily determined. This suggests an inherent sparseness in the human neuro-visual representation of the object. Therefore, studying such sparse sketches can aid our understanding of the cognitive processes involved and spur the design of efficient visual classifiers. 

\begin{figure*}[ht!]
  \includegraphics[height=3.5cm,width=\textwidth]{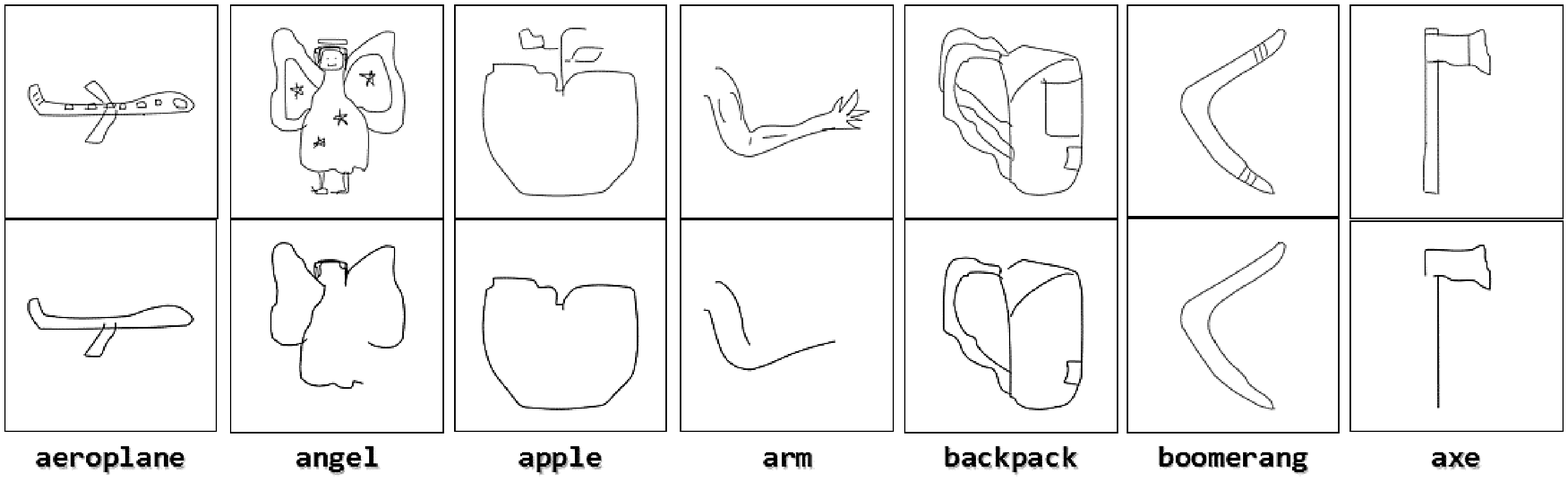}
  \caption{Original sketches (top row) and corresponding \textit{category-epitome}s (bottom row) for various object categories.}
	\label{fig:primal-sketches}
\end{figure*}

Freehand line sketches are typically formed as a composition of primitive hand-drawn curves (called strokes) added sequentially over time. A significant body of work has examined such sketches in the context of classification and content-based retrieval problems\cite{Hu:2013:PEG:2479988.2480107}\cite{Hu2010}\cite{KMY06}\cite{QiGLZXS13}. In these problems, the end product of the sketching process, i.e. the full sketch, is typically considered \textit{in-toto}.	 However, we believe it can be quite instructive to study the temporal process of sketch formation itself, starting with the first hand-drawn stroke until the last stroke which finalizes the sketch. Our belief originates in a surprising discovery we have made : For a given sketch, there exists a minimal discriminative subset of all its strokes which contribute to the sketch's identity (category) being recognized consistently and correctly. We term the sketch composed using this minimal stroke subset as a \textit{category-epitome}. Figure \ref{fig:primal-sketches} shows examples of freehand line sketches and their corresponding sparse \textit{category-epitome}s.

In this paper, we describe how these sparse \textit{category-epitome}s are obtained. The \textit{category-epitome} has a unique feature which sets it apart from other methods of sparse sketch generation: The process of epitome construction guarantees that the most fundamental strokes which enable category recognition (discrimination) are retained. To quantify the sparseness of an epitome, we provide a natural measure termed epitome-score. We analyze the eptiomes and corresponding epitome-scores for freehand sketches across various object categories.Our analysis provides a  novel viewpoint for studying the semantic nature of object categories. 

The rest of the paper is organized as follows: We briefly review related literature in Section \ref{sec:relatedwork}. In Section \ref{sec:sketchclassifier}, we describe the construction of a sketch classifier. This sketch classifier plays a crucial role in obtaining \textit{category-epitome} of a given sketch. In Section \ref{sec:catepitome}, we describe how the \textit{category-epitome} of a sketch is actually obtained and provide a simple, natural measure termed epitome-score to quantify its epitomal-ness. Section \ref{sec:analysis} contains an analysis of \textit{category-epitomes} and epitome-scores across object categories. Section \ref{sec:conclusion} concludes the paper by outlining some of the promising directions for future work.


\section{Related Work}
\label{sec:relatedwork}

In his seminal work \cite{Marr}, Marr suggested an abstract representation called the \textit{primal sketch} -- a sparse, sketch-like representation of generic images in terms of image primitives. Inspired by his theories, methods for formalizing the notion of primal sketch have been proposed \cite{Guo:2003:TMT:946247.946727} and primal sketch representations have been used as features for object detection\cite{KMY06}, texture characterization\cite{haralick_topographic_1983} and for super-resolution\cite{Sun03imagehallucination}.
Yet another line of research aims to generate sketches from one or more source images\cite{QiGLZXS13}\cite{MarvaniyaBMM12} without explicit recourse to the idea of primal sketch. A common feature in all these works is the utilization of photographic images as the starting point. In contrast, our starting point is the sketch stroke data created by human beings. This provides a glimmer of hope that the sparse neurovisual representation of the object being sketched is transferred to the sketch in the process of drawing, at least in part. 

In addition to the artificial (i.e. not generated by human hand) nature of sketch generation, the works mentioned above do not attempt to quantify the sparseness of the resulting sketch. They also do not examine the temporal nature of sketch composition. In contrast, recent work by Berger et. al.\cite{Berger:2013:SAP:2461912.2461964} analyzes the temporal aspect of sketching in the context of mimicking artist style. However, their emphasis is on synthesizing abstract facial sketches rather than recognition. Moreover, their sketches are produced by professional artists. In contrast, our sketches have been generated by crowdsourcing from the general public. The idea of identifying and utilizing a discriminative subset of strokes was employed by Karteek et al.\cite{Alahari:2005:DSO:1106779.1106963} for classifying online handwritten data. Another work close in spirit to ours is the unpublished, but publicly available manuscript of  Jiang et al.\cite{jiangzhu} which examines the temporal evolution of sketches from a visualization perspective. Finally, the work of Eitz et al.\cite{eitz} examines how humans tend to draw objects by analyzing a large number of sketches spread across commonly encountered object categories. We employ their database of sketches in this paper. 

\section{Building the sketch classifier}
\label{sec:sketchclassifier}

\begin{figure*}[ht]
	\centering
		\includegraphics[height=2.5cm,width=1.00\textwidth]{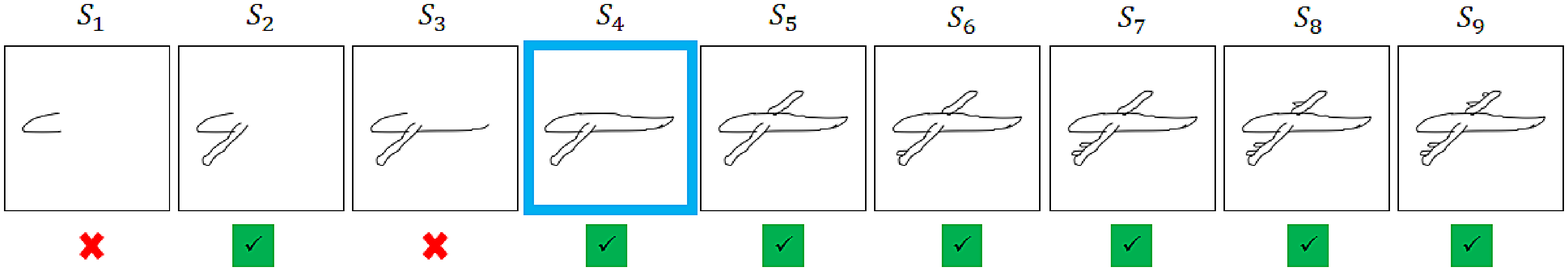}
	\caption{Constructing the \textit{category-epitome} for an \texttt{airplane} sketch: The sketch has 9 strokes. $S_1 - S_9$ are the cumulative stroke sequence canvases. A red cross mark indicates a misclassification($0$) while a green tick mark indicates correct classification($1$). Canvas $S_4$ outlined by a cyan rectangle is the \textit{category-epitome}. Note that even though canvas $S_2$ is classified correctly, we consider \textit{category-epitome} as canvas $S_4$, the temporally earliest, correctly classified canvas whose successors are classified correctly as well.}
 \label{fig:computing-primality}
\end{figure*}

\subsection{The sketch database}
The sketches used in our study have been taken from the publicly available freehand line sketch database of Eitz et al.\cite{eitz}. This database contains a curated set of $20,000$ hand-drawn sketches evenly distributed across $250$ object categories. As mentioned before, these sketches have been obtained by crowdsourcing across the general population. As such, they are a good starting point for analyzing the underpinnings of the sketching process by humans. A few examples from the database can be seen in the top row of Figure \ref{fig:primal-sketches}. The dominant appeal of this database is that the temporal stroke information (the sequential order in which the strokes were drawn) for a sketch has been provided. As we will see in Section \ref{sec:catepitome}, it is precisely the temporal stroke information that forms the basis for obtaining the \textit{category-epitome} and the corresponding epitome-score for sketches. 

\subsection{Sketch data augmentation}
\label{sec:augmentation}

The database of Eitz. et. al. contains $80$ sketches per category. To increase the number of sketches per category available for training the classifier (Section \ref{sec:classif}), we perform data augmentation by applying geometric and morphological transformations to each sketch. Specifically, each sketch is initially subjected to image dilation using a $5 \times 5$ square structuring element. A number of transforms are applied to this thickened sketch --  mirroring (across vertical axis), rotation ($\pm 5,\pm 15$ degrees), combinations of horizontal and vertical shifts ($\pm 5,\pm 15$ pixels), central zoom ($\pm 3\%,\pm 7 \%$ of image height). As a result, $30$ new sketches are generated per original sketch. The data augmentation procedure results in $2400$ sketches per category, for a total of $600,000$ sketches across $250$ categories.

\subsection{Sketch feature extraction}
\label{sec:sketchfeatex}

As the top row of Figure \ref{fig:primal-sketches} demonstrates, the spatial density of sketch strokes can be quite small. Moreover, extracting typical image features (e.g. based on texture) is ruled out and edge information is quite sparse. Nevertheless, a number of alternatives have been proposed as sketch features. For a survey of these methods in the context of sketch-based image retrieval, refer to \cite{Hu:2013:PEG:2479988.2480107}. We extract Histogram of Oriented Gradients(HOG\cite{HOG})-like sketch descriptors using the pipeline described by Eitz et. al.\cite{eitz}. The collection of descriptors so obtained are then combined using Fisher image representation approach\cite{perronnin2007fisher} to obtain a feature vector for each sketch.

\subsection{Sketch Classification}
\label{sec:classif}
As an initial exploration and for ease of analysis, we consider only the first $50$ alphabetically sorted sketch categories. We build a sketch classifier by utilizing $80\%$ of the augmented sketches(Section \ref{sec:augmentation}) from each category for training and the rest for testing. Doing so provides $2400 \times 0.8 \times 50 = 96,000$ sketches for training and $80 \times 0.2 \times 50 = 800$ sketches for testing\footnote{Testing is done only on the original sketch subjected to dilation and not on its subsequently transformed variants generated for data augmentation. Hence the factor of $80$ for each test category instead of the full $2400$ as in training.}.

From each test sketch, its Fisher feature vector(Section \ref{sec:sketchfeatex}) is obtained.  A multi-class Support Vector Machine (SVM) classifier employing a Radial Basis Function (RBF) kernel was trained on the Fisher feature vectors by employing $5$-fold cross validation and grid-based parameter search. The accuracy of the resulting classifier was $60.25\%$. For context, the accuracy (for the same split ratio of training and test sketches) obtained by Eitz et. al. \cite{eitz} is $54\%$ with the caveat that the number of categories are larger than ours -- $250$.

In the overall scheme of things, the sketch classifier is only useful to the extent that it lets us determine the \textit{category-epitome}s. Seen in this light, the accuracy of the classifier assumes secondary importance. However, a classifier with good performance is still desirable as it ensures a larger coverage of the test set. In addition, such a classifier could potentially help obtain sparser \textit{category-epitome}s compared to a counterpart whose performance is relatively poor. Nevertheless, to progress towards our goal of determining \textit{category-epitome}s, we settle for the existing performance of the sketch classifier. In the next section, we shall see how the sketch classifier is actually utilized in constructing the \textit{category-epitome}.

\section{Obtaining the category-epitome}
\label{sec:catepitome}	

\subsection{Constructing Cumulative Stroke Sequences}
\label{sec:sketch-sequence-construction}

As the first step in determining the \textit{category-epitome}, we construct sequences of cumulative strokes derived from correctly classified test sketches. Suppose the sequence of strokes in the temporal order they were drawn in a test sketch is given by $S = \{s_1,s_2\ldots s_N\}$ where $N$ is the total number of strokes in the sketch. To construct the corresponding cumulative stroke sequence, we begin with a blank canvas. Strokes from the given sketch are successively added to the blank canvas in the temporal order. As each stroke is added, intermediate canvases $S_1,S_2 ,\ldots ,S_N $ are created. Specifically, the intermediate canvases are given by $S_1 = \{s_1\},S_2=\{s_1,s_2\},\ldots ,S_N=\{s_1,s_2,\ldots ,s_N\}$. At the end of this process, we obtain the cumulative stroke sequence $CSS = \{S_1, \ldots ,S_N\}$. Figure \ref{fig:computing-primality} illustrates the creation of cumulative stroke sequence for a sketch from \texttt{airplane} category.

\subsection{Constructing the category-eptiome}

Having generated the cumulative stroke sequence for a sketch as described above, the \textit{category-epitome} can be constructed. Using the classifier from Section \ref{sec:sketchclassifier}, each intermediate canvas of the cumulative stroke sequence is classified to obtain a binary labeling -- the label is $1$ if the sketch category is correctly identified and $0$ otherwise. Thus, we obtain a binary label sequence $\mathcal{L} = \{l_1,l_2,\ldots ,l_N\}$ corresponding to each canvas of the cumulative stroke sequence $CSS$(see Figure \ref{fig:computing-primality}). 
 
Note that the final canvas $S_N$ corresponds to the original test sketch since all the strokes have been added to the canvas at that point. Therefore, the final classification label $l_N$ must be $1$ since we are working with correctly classified test sketches. Now, consider the product sequence $\mathcal{P} = \{P_1,P_2..\ldots P_N\}$ formed by cumulative multiplication of labels $l_i \in \mathcal{L},i=1,2\ldots N$:

\begin{align}  
\label{seq}
P_i = \prod_{j=i}^{N} l_j
\end{align}

Then, the \textit{category-epitome} corresponds to canvas $S_e$ of the cumulative stroke sequence such that

\begin{align}  
\label{primality}
 e = \min_{1 \le i \le N} \{i | P_i = 1 \}
\end{align}

Informally, the \textit{category-epitome} $S_e$ is the temporally earliest, correctly classified canvas whose successors are classified correctly as well. Using the example in Figure \ref{fig:computing-primality}, the classification label sequence is given by $\mathcal{L} = \{0,1,0,1,1,1,1,1,1\}$. From Equation \eqref{seq}, the product sequence is computed as $\mathcal{P} = \{0,0,0,1,1,1,1,1,1\}$. From Equation \eqref{primality}, we obtain $e=4$. In other words, canvas $S_4$ (outlined by a cyan rectangle in Figure \ref{fig:computing-primality}) corresponds to the \textit{category-epitome}: the temporally earliest, correctly classified canvas whose successors $S_5 \ldots S_9$ are classified correctly as well. Figure \ref{fig:primal-sketches} shows sketches from various categories and their corresponding \textit{category-epitome}s. 

\subsection{Epitome-score : Quantifying the \textit{category-epitome}}

Our procedure for obtaining the \textit{category-epitome}, described above, also provides a natural method for quantifying the ``epitome"-ness of the original, full sketch. Using $e$ obtained from Equation \eqref{primality}, we define the epitome-score $\mathcal{E}$ of a sketch as :

\begin{align}
\label{episcore}
\mathcal{E} = 
\begin{cases}
\frac{e}{N},\mbox{$e \neq 1$} \\
0,\hspace{1.5mm} \mbox{$e=1$}
\end{cases}
\end{align} 

where $N$ is the total number of strokes in the sketch. $e=1$ corresponds to the situation where merely drawing the first stroke conveys the epitome-ness of the sketch. Therefore, we have defined the corresponding epitome-score $\mathcal{E}$ to be $0$ for consistency across sketches. Our definition of epitome-score  $\mathcal{E}$ essentially conveys the sparseness underlying the sketch -- the smaller its value, the more sparser the sketch is likely to be. Epitome-scores very close to $1$, on the other hand, indicate that very few strokes in the original sketch are redundant. Fortunately, very high epitome-scores are not the norm, as we will see -- sparsity is pervasive across categories. Referring once again to Figure \ref{fig:computing-primality}, the epitome-score for the \texttt{airplane} sketch is computed as $\mathcal{E} = \frac{4}{9} = 0.44$. 

\section{Analysis}
\label{sec:analysis}

\begin{figure*}[!ht]        
        \centering      	
		\includegraphics[width=1.0\linewidth, height=0.3\textwidth]{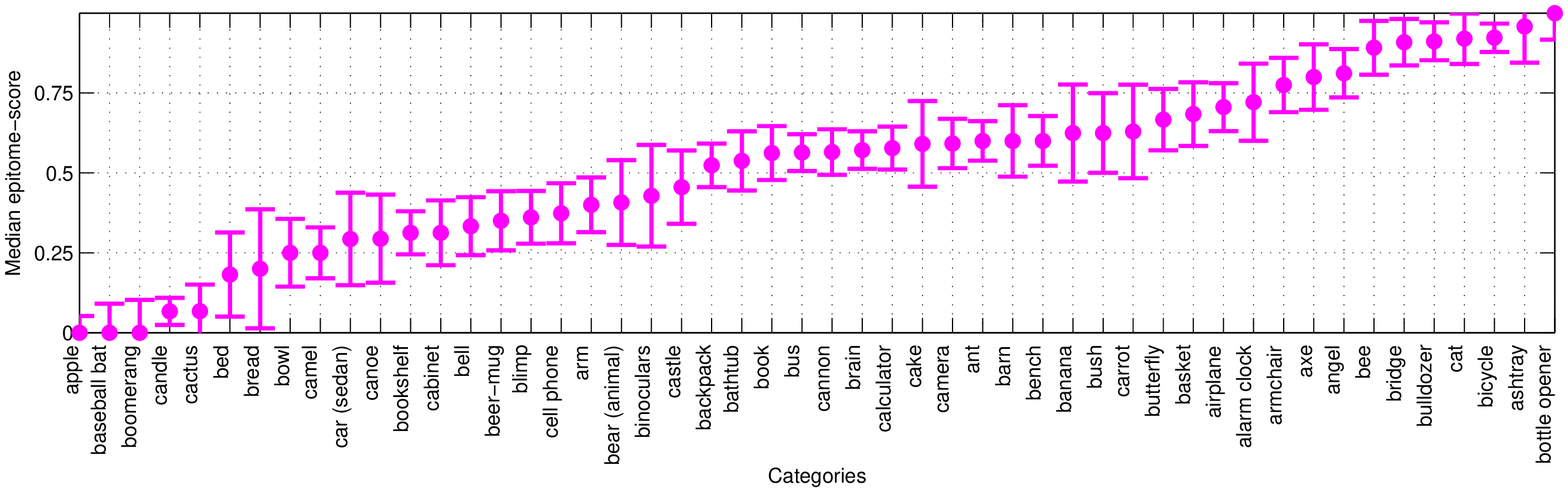}	
		 \caption{Median epitome-scores (y-axis) and corresponding error bars for $50$ object categories(x-axis). The  standard errors are clamped to $[0,1]$ -- the range of epitome-scores. The median scores are shown as filled circles. }	
	\label{fig:primality-with-errorbars}
	\centering	 
\end{figure*}
\begin{figure}
\centering
	\includegraphics[width=0.49\textwidth]{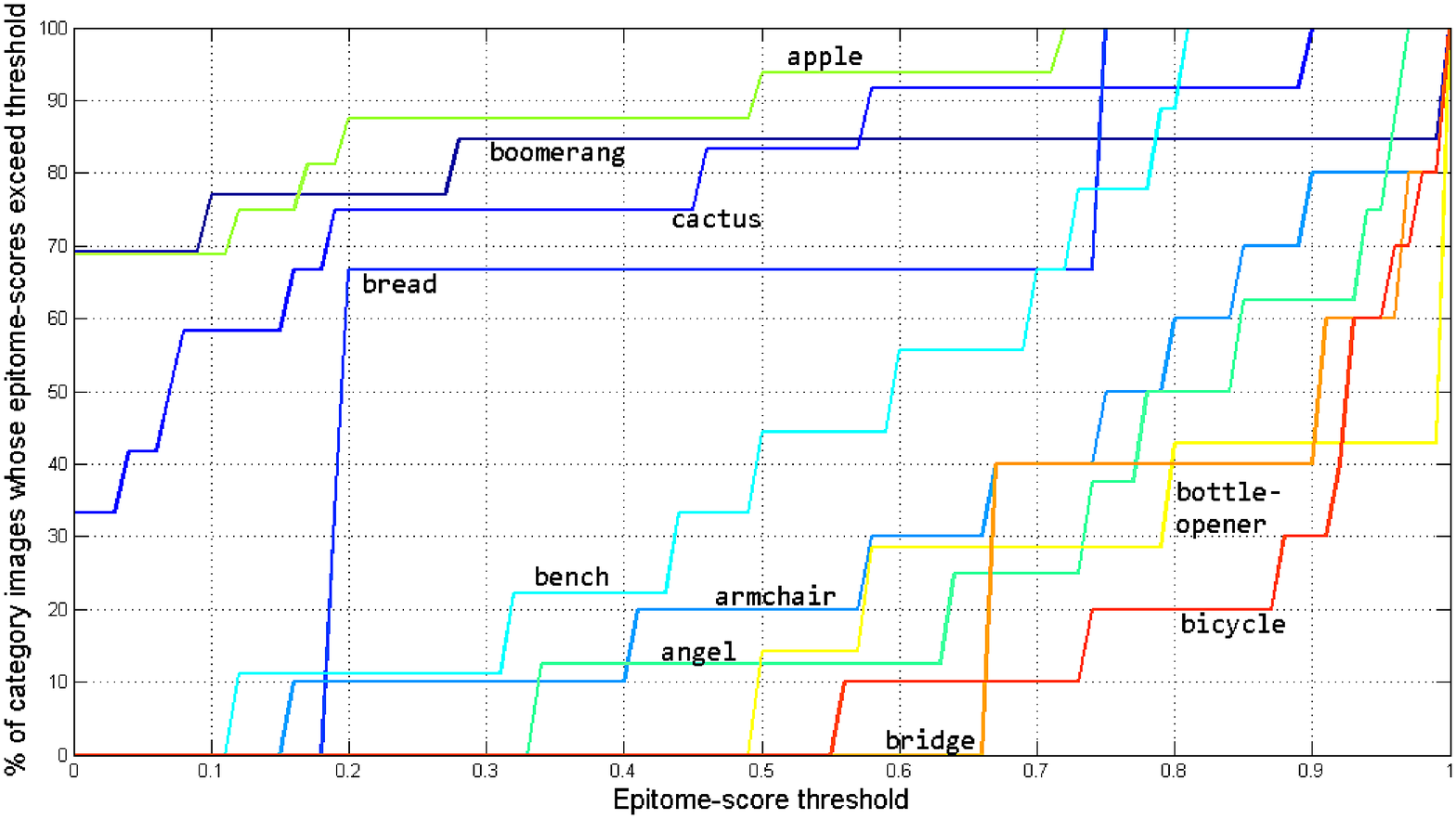}
	\caption{Demonstrating the effect of budgeting epitome-score on category recognition rate using data from $10$ selected categories. Values of various possible epitome-score thresholds are shown on x-axis. The $\%$ of images in each category which exceed a particular value of the threshold are shown on y-axis. The $10$ category names are shown adjacent to their respective plots.}
	\label{fig:primality-trend-length}	
 \end{figure}

\subsection{Epitome-scores across categories}
\label{sec:primality-across-categories}

We begin our analysis by computing the median of epitome-scores for sketches on a category-by-category basis. Figure \ref{fig:primality-with-errorbars} displays the median epitome-scores (shown as filled circles in the figure) for test sketches across $50$ object categories. It is heartening to observe that $42\%(21/50)$ of the categories have median epitome-scores below $0.5$ and $80\%(40/50)$ of the categories have scores below $0.75$. These trends suggest an inherent sparseness for visual representation of object categories because the smaller the epitome-score, the sparser the \textit{category-epitome} sketch. If we examine the scores closely, the epitome-scores for some of the categories (\texttt{apple} and \texttt{boomerang}) are $0$. For these categories, the sketches (see Figure \ref{fig:primal-sketches}) are typically drawn such that a dominant stroke  (an oval in the vertical plane with a notch at the top for the apple and the chevron-like contour for the boomerang) essentially captures the epitome-ness of the category(See also Equation \ref{episcore}). 

The varying lengths of error bars in Figure \ref{fig:primality-with-errorbars} indicates that some object categories may have multiple representative \textit{category-epitome}s rather than a single, unique representative. For example, the sketches of the category (human) \texttt{arm} are likely drawn with a fairly consistent appearance by humans. This consistency influences the sketch classifier and can cause it to produce \textit{category-epitome}s which exhibit a minor amount of variations, thus resulting in a compact distribution of epitome-scores (shorter error bars). In contrast, the variety in sketches of a category such as \texttt{carrot} is reflected in the corresponding \textit{category-epitome}s and by extension, in the longer error bars of its epitome-scores.
\subsection{Epitome-score as a proxy for semantic level of detail}
\label{sec:primality-as-level-of-detail}

A different perspective can be gained by examining categories for a given value of epitome-score. For each category, we count the number of test sketches whose epitome-score exceeds a threshold and normalize by the number of test sketches in the category. To facilitate analysis and avoid visual clutter, we select $4$ prototypical categories whose plot contours occupy the extremities(towards top-left corner and bottom-right corner) of the original plot. In addition, we select $6$ other prototypical categories whose plot contours lie between the extremity plots previously mentioned. The resulting plot can be viewed in Figure \ref{fig:primality-trend-length}. 

Epitome-score can be considered as a proxy for semantic level-of-detail. Viewed in this light, the plot from Figure \ref{fig:primality-trend-length} suggests the varying epitome-score budgets across categories -- some categories require a considerable level of detail before their epitomal avatars are revealed (e.g. categories with plots towards the lower right such as \texttt{bridge} and \texttt{bicycle}). On the other hand, categories with plots towards the upper left corner (\texttt{apple}, \texttt{boomerang}) have relatively less stringent demands on level of detail. 

To view a larger set of \textit{category-epitome}s,  t-SNE\cite{tsne} visualizations of epitomes for select categories and additional results which support the analysis presented in this section, visit \url{http://val.serc.iisc.ernet.in/sketchepitome/se.html} \hspace{1.5mm}.

\section{FUTURE WORK}
\label{sec:conclusion}
A number of directions exist for future work. One obvious direction would be to improve the performance of existing sketch classifier in terms of number of categories as well as accuracy. A well-performing classifier which utilizes fewer training samples translates to a potentially larger set of test \textit{category-epitomes} for analysis. We also intend to compare our current distribution of \textit{category-epitome}s/epitome-scores and related analysis with that obtained when human subjects are asked to identify the cumulative stroke sequences. For this comparison, we plan to use the benchmark sketch database created by Ros\'{a}lia et al.\cite{rosalia} who employ a human-evaluation based technique to identify a subset of $160$ non-ambiguous\footnote{Ros\'{a}lia et al.\cite{rosalia} define a non-ambiguous sketch as one whose identity is agreed upon by at least $2$ people among the subjects surveyed.} object categories from the $250$ originally provided by Eitz et al\cite{eitz}. However, as the number of categories increases, visualizing trends (Figures \ref{fig:primality-with-errorbars} - \ref{fig:primality-trend-length}) becomes a challenge. Therefore, novel visualization methods need to be explored or alternately, the number of categories needs to be curated in a meaningful manner for representative sketch analysis.

\bibliographystyle{IEEEbib}
\bibliography{refs}

\end{document}